\documentclass[letterpaper, 10 pt, conference]{ieeeconf}  

\IEEEoverridecommandlockouts                              

\usepackage{tabularx}
\usepackage{graphicx}
\usepackage{xcolor}

\title{\LARGE \bf
From Monocular Vision to Autonomous Action:\\Guiding Tumor Resection via 3D Reconstruction
}

\author{
Ayberk Acar$^{1+}$, Mariana Smith$^{2+}$, Lidia Al-Zogbi$^{1}$, Tanner Watts$^{3}$, Fangjie Li$^{1}$, Hao Li$^{1}$,\\
Nural Yilmaz$^{2}$, Paul Maria Scheikl$^{2}$, Jesse F. d'Almeida$^{4}$, Susheela Sharma$^{4}$, Lauren Branscombe$^{5}$,\\ Tayfun Efe Ertop$^{6}$,
Robert J. Webster III$^{4}$, Ipek Oguz$^{1}$, Alan Kuntz$^{3}$, Axel Krieger$^{2}$, Jie Ying Wu$^{1}$
\thanks{+Both authors contributed equally.}
\thanks{This material is supported in part by the Advanced Research Projects Agency for Health (ARPA-H) under grant number D24AC00415, and by the NSF Foundational Research in Robotics (FRR) Faculty Early Career Development Program (CAREER) under grant number 2144348. }
\thanks{$^{1}$Department of Computer Science, Vanderbilt University, Nashville, TN 37235, USA
        {\tt\small ayberk.acar@vanderbilt.edu}}%
\thanks{$^{2}$Department of Mechanical Engineering, Johns Hopkins University, Baltimore, MD 21211, USA
        {\tt\small msmit458@jh.edu}}%
\thanks{$^{3}$Robotics Center and Kahlert School of Computing, University of Utah, Salt Lake City, UT 84112, USA}
\thanks{$^{4}$Department of Mechanical Engineering, Vanderbilt University, Nashville, TN 37235, USA}
\thanks{$^{5}$Virtuoso Surgical, Nashville, TN 37205, USA}
\thanks{$^{6}$Department of Mechanical, Aerospace and Biomedical Engineering, University of Tennessee, Knoxville, TN 37996, USA}
}

\begin{document}

\maketitle
\thispagestyle{empty}
\pagestyle{empty}

\begin{abstract}
Surgical automation requires precise guidance and understanding of the scene. Current methods in the literature rely on bulky depth cameras to create maps of the anatomy, however this does not translate well to space-limited clinical applications. Monocular cameras are small and allow minimally invasive surgeries in tight spaces but additional processing is required to generate 3D scene understanding. We propose a 3D mapping pipeline that uses only RGB images to create segmented point clouds of the target anatomy. To ensure the most precise reconstruction, we compare different structure from motion algorithms' performance on mapping the central airway obstructions, and test the pipeline on a downstream task of tumor resection. In several metrics, including post-procedure tissue model evaluation, our pipeline performs comparably to RGB-D cameras and, in some cases, even surpasses their performance. These promising results demonstrate that automation guidance can be achieved in minimally invasive procedures with monocular cameras. This study is a step toward the complete autonomy of surgical robots.

\end{abstract}

\section{Introduction}
\label{introduction}
Surgical automation has the potential to improve clinical outcomes, standardize operational workflows, and increase efficiency and safety~\cite{han2022systematic}. Visual guidance is essential for autonomous robotic surgeries, enabling precise, and reliable performance during procedures. Automated procedures have used modalities such as RGB-D~\cite{hwang2020applying, kim2024towards} or stereo cameras~\cite{kehoe2014autonomous} for scene understanding and mapping. However, compared to monocular counterparts, these cameras have higher costs and their use is limited in space restricted environments like minimally invasive surgeries. As an alternative, 3D reconstruction pipelines using a series of monocular RGB images can be used for in-vivo environment mapping~\cite{zhong2025review}. Although these monocular reconstruction methods are promising, evaluation of their use in a downstream task is limited. 

\begin{figure}[tbh]
\centering
\includegraphics[trim={0cm 0cm 0cm 0cm},clip,width=0.85\columnwidth]{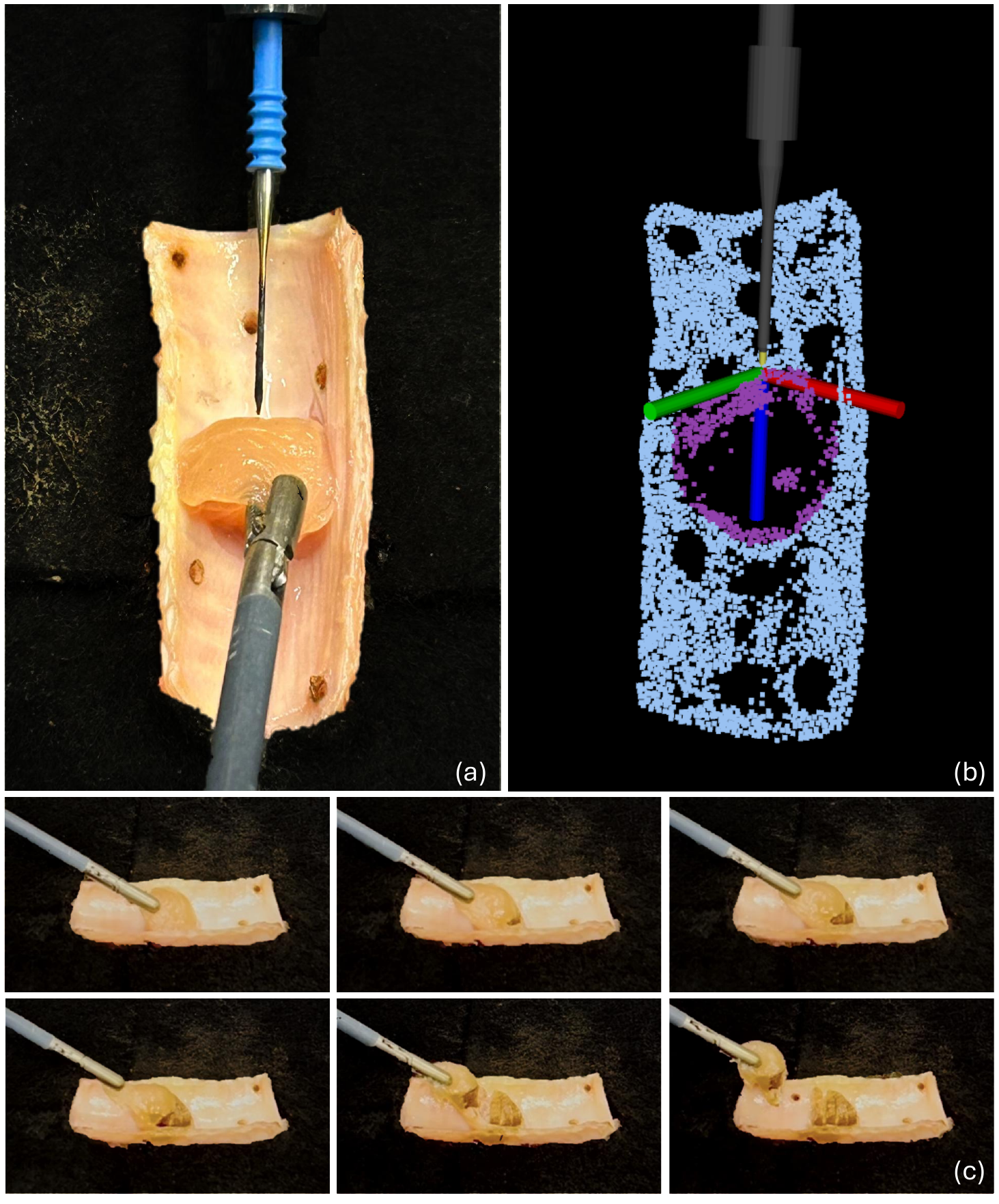}
\caption{Robotic electrocautery tool positioned on an open-surgery CAO tissue model (a), corresponding RViz visualization of the segmented point cloud generated by SfM reconstruction and registration (b). Snapshots taken during the robotic CAO resection of this tissue model using the SfM (c). }
\label{IntroFig}
\end{figure}
In this study, we propose a visual guidance pipeline for automated central airway obstruction (CAO) surgery using monocular images. CAO refers to blockage of the main airways \cite{ernst2004central}, which affects more than 80,000 patients annually \cite{chen1998malignant}. This procedure is traditionally performed manually using a bronchoscope \cite{ayers2001rigid, semaan2015rigid}, though minimally invasive robotic systems now offer alternatives that can reduce procedural trauma \cite{gafford2020concentric}. 

We test the feasibility of the proposed pipeline as a guidance method on a tumor removal task. Our contributions are 1) a comprehensive evaluation of structure from motion (SfM) algorithms on an ex-vivo CAO dataset as a scene mapping tool, 2) method for creating 3D segmented maps using monocular images for trachea and tumor 3) an open surgery demonstration of tumor removal using SfM reconstruction of the surgical scene (Fig.\ref{IntroFig}). This work is a step toward a fully-autonomous CAO surgery by providing scene understanding with a standard monocular endoscope. The pipeline proposed here has the potential to extend to other robotic minimally invasive surgical procedures.

\section{Background}
\label{Background}

\subsection{Surgical Autonomy}
Previous studies have demonstrated automated tumor resection across various organs; however, these procedures require additional information beyond RGB images. Ge \textit{et al.} and Saeidi \textit{et al.} showed tumor removal examples for squamous cell carcinoma, with autonomous systems using near-infrared fluorescent markers as aid to their visual pipelines~\cite{ge2021supervised,ge2023autonomous,saeidi2019supervised}. Hu \textit{et al.} used multimodal scanning fiber endoscope images and keypoint matching to create a 3D reconstruction of the scene, and demonstrated a semi-autonomous brain tumor removal in a simulated laboratory environment~\cite{hu2018semi}. While effective, adding markers and relying on fluorescence adds complexity to a surgical workflow. Smith \textit{et al.} presented a supervised autonomous workflow for CAO resection using an RGB-D camera, focusing particularly on cut trajectory planning \cite{jmrr}. The RGB-D camera limits applicability in minimally invasive surgeries. 

\subsection{Monocular Mapping}
3D reconstruction algorithms based on RGB images offer a cost and space-effective alternative to RGB-D cameras. SfM algorithms mainly aim to provide accurate offline 3D reconstructions of the scene using 2D images~\cite{zhong2025review}. Methods such as COLMAP~\cite{schoenberger2016sfm} and hierarchical localization toolkit (hloc)~\cite{sarlin2019coarse} offer a range of features for 3D reconstruction from unordered images. These pipelines are used for endoscopy as well, in combination with other methods or as comparison baselines~\cite{acar2024towards,liu2020extremely,liu2020reconstructing}. With the emergence of new keypoint detection and matching methods, these pipelines are updated and their performances require re-evaluation.

To the best of the authors' knowledge, no prior work has reported a fully integrated robotic system utilizing 3D reconstruction from regular 2D images to guide autonomous surgical interventions. In this manuscript, we integrate our proposed monocular SfM approach into the workflow from \cite{jmrr}, aiming to eliminate reliance on the RGB-D data stream. 

 


Our technical pipeline comprises several distinct components. We explain each component and the experiments to validate them in Section~\ref{MethodsandExperiments}. Outcomes of each experiment are given in Section~\ref{Results}. We discuss the results of experiments in Section~\ref{Discussion}, and conclude in Section~\ref{Conclusion}.

\section{Methods and Experiments}
\label{MethodsandExperiments}

We begin by introducing our pipeline for the enclosed CAO model, which simulates surgical imagery. Next, to assess the ability of segmented reconstructions to guide automation, we adapt the pipeline to an open model and explain the downstream tumor resection task along with its evaluation tests.

\subsection{CAO Model Reconstruction} 

\subsubsection{Data Preparation}\label{DataPreparation} To create a dataset resembling endoscopic images from CAO surgeries, we prepared five ex-vivo models using sheep pluck and chicken breast tissue. Chicken breast was cut to approximate tumors causing about 50\% tracheal occlusion, and subsequently affixed to the airway surface with superglue. Using the Virtuoso Surgical (Nashville, TN, USA) endoscopy system, we recorded a total of 9 video sequences of varying lengths of 276 to 1747 frames, under different lighting conditions. We acquired a total of 8554 images and cropped out the circular endoscope mask by a center crop of $545\times545$ pixels.

\subsubsection{Structure from Motion Algorithm Selection} 
We comprehensively evaluate six different state-of-the-art SfM pipelines,  with different feature detectors and matchers, from the hloc toolkit \cite{sarlin2019coarse} for reconstructing the five ex-vivo CAO models described above: SuperPoint~\cite{detone2018superpoint} + SuperGlue~\cite{sarlin2020superglue}, SuperPoint + LightGlue~\cite{lindenberger2023lightglue}, SuperPointInLoc~\cite{taira2018inloc} + LightGlue, ALIKED~\cite{zhao2023aliked} + LightGlue, SIFT~\cite{lowe1999object} NN, and DISK~\cite{tyszkiewicz2020disk} + LightGlue. The reconstruction process employs NetVLAD \cite{arandjelovic2016netvlad} for estimating covisible image pairs via image retrieval. To ensure reasonable processing times, the maximum number of matches between image pairs was limited to 50. All reconstructions were executed on an NVIDIA RTX4090 GPU. We evaluate the pipelines' performance using standard SfM metrics: percentage of successfully reconstructed images, average track length, total number of observations, mean reprojection error, and computational runtime. The best-performing SfM pipeline is chosen for all subsequent experiments.

\subsubsection{Segmented Point Cloud Generation}\label{Sec:SegmentedPC}
The segmentations are generated following the pseudo-label learning strategy~\cite{liu2024cosst} through three steps: (1) Generating initial pseudo-labels in a zero-shot manner using Segment Anything Model 2 (SAM2)~\cite{ravi2024sam}; (2) Using generated pseudo-label for self-training with SAM1~\cite{kirillov2023segment} and (3) refining the segmentation results from step (2) using morphological operations to remove false positives, followed by an additional round of self-training. We use the SAM1 architecture as the final segmentation model to ensure consistency in comparing downstream task results with previous work~\cite{jmrr}.

To assign labels to each point in the reconstruction, the 3D point cloud is projected onto the segmentation masks using the radial camera model and parameters estimated by the selected SfM algorithm. We implement a weighted voting approach for points visible across multiple images, based on the principle that labels closer to the image center are more reliable due to reduced effects from occlusion, illumination variations, and distortion-related projection errors. The weight $\mathcal{W}$ is calculated according to Equation (\ref{weightequation}), where $d$ is the Euclidean distance between the projected pixel and image center: 
\begin{equation}
\label{weightequation}
    \mathcal{W} = 1/(1+d).
\end{equation}

\begin{figure}[tbp]
\centering
\includegraphics[trim={0cm 0cm 0cm 0cm},clip,width=0.75\columnwidth]{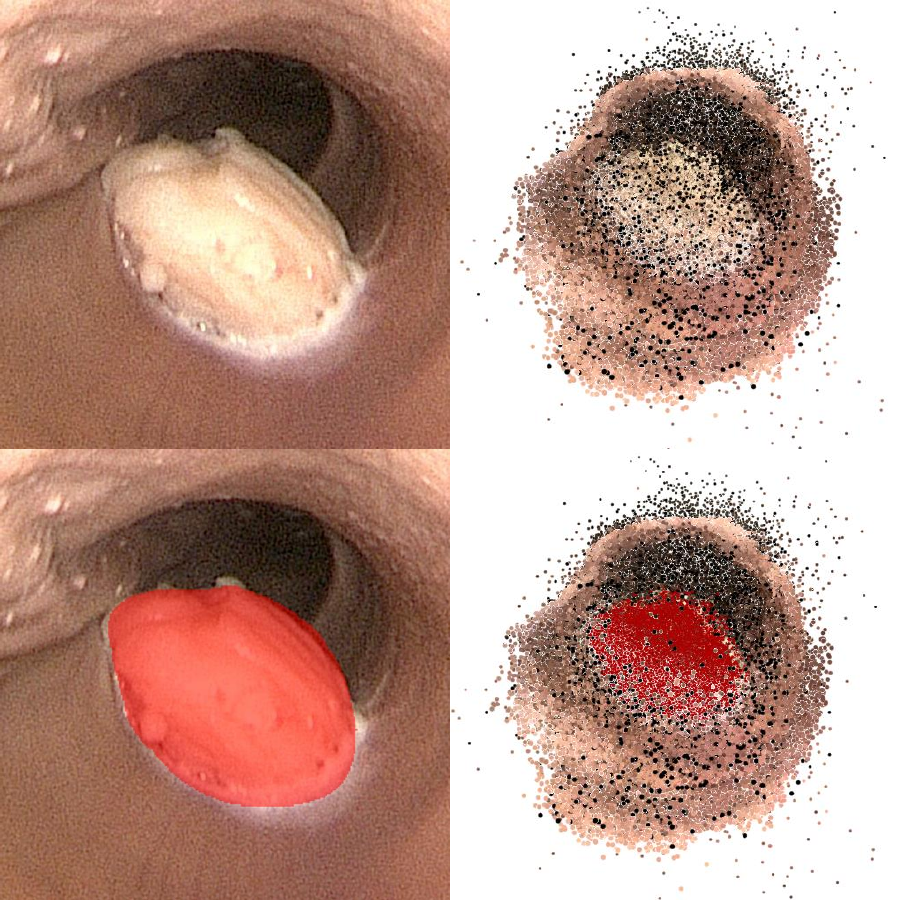}
\caption{Example images (left) and corresponding reconstructions (right) from the CAO dataset. The upper row shows unsegmented reconstructions, while the lower row shows the same model with segmentation applied.}
\label{SegmentationPC}
\end{figure}

This approach produces segmented point clouds that delineate tumor boundaries within the reconstructed model (Fig.~\ref{SegmentationPC}).

To analyze the segmentation consistency and label projection accuracy, we manually segment 20 images in a video sequence that the segmentation model is not trained on. These 20 images correspond to different viewpoints of the reconstruction, and we reproject the final segmented point cloud to the manual segmentation masks. The percentage of 3D points labeled as tumors inside the 2D tumor segmentation masks gives the precision of point cloud segmentation.

\subsection{Autonomous Robotic Resection} 

To assess the feasibility of using our proposed segmented point cloud reconstruction pipeline for robotic guidance, we evaluate its performance in a tumor resection task. This study maintains the same methodology developed in \cite{jmrr}, but substitutes the original RGB-D approach with our monocular reconstruction pipeline.  Since the resection task in \cite{jmrr} used a large UR robot, with a bulky RGB-D camera as a placeholder for SfM, it required an open surgery approach on open tissue models. These open tissue models were made from chicken breast tumors and 'half-pipes' of ex vivo porcine trachea. To enable our comparison study, we evaluate our SfM approach on open tissue models as in \cite{jmrr}.

\begin{figure}[t]
\includegraphics[trim={0cm 0cm 0cm 0cm},clip,width=0.95\columnwidth]{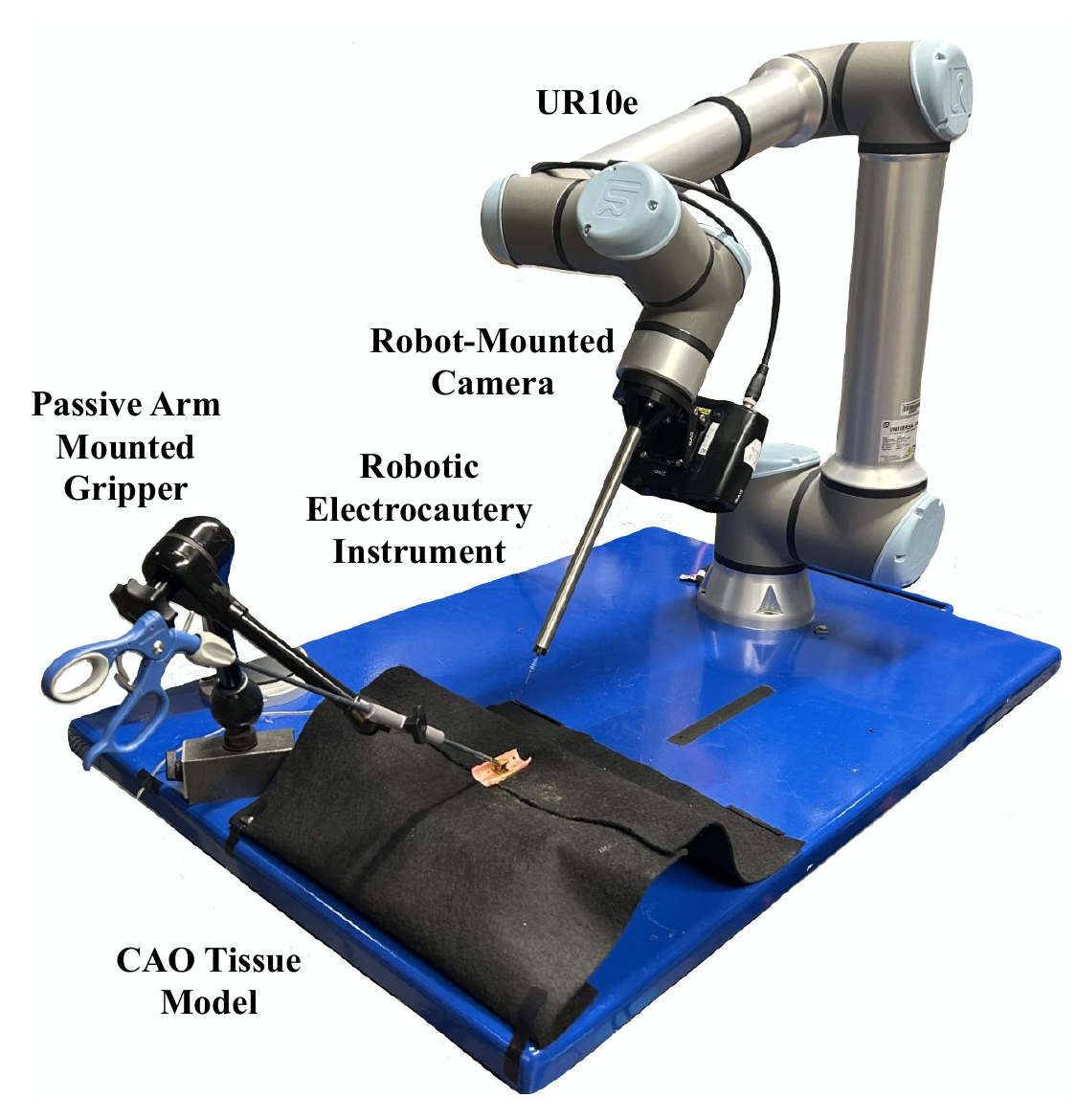}
\caption{Full robotic system for downstream CAO resection experiments.}
\label{SystemFig}
\end{figure}
\subsubsection{Testbed}
The system comprises a UR10e robot (Universal Robots, Odense, Denmark) equipped with a Zivid 2 Plus M60 camera (Oslo, Norway), and a custom electrocautery instrument using a monopolar electrode (Bovie, Clearwater, FL), as illustrated in Fig. \ref{SystemFig}. A laparoscopic gripper (Snowden-Pencer SP90-6379, Tucker, GA) attached to a passive hand-tightened NOGA arm (Noga Engineering, Shlomi, Israel) enables controlled manual translation along its primary axis between successive cuts. Due to the system's size constraints, the methodology from \cite{jmrr} can only be applied to ``half-pipe" preparations of ex-vivo tracheas. Accordingly, five open tissue models were created as described in Section~\ref{DataPreparation}, but with porcine tracheas cut longitudinally in half. Each trachea specimen was secured to an electrosurgical grounding plate using four stay sutures. The grounding plate was covered with a black cloth to prevent unwanted reflections in the imaging field.

\subsubsection{Open CAO Reconstruction}
RGB images for reconstruction were captured from 20 different camera positions by moving the camera around 10-20 cm between each image, covering an area of approximately 0.2-0.4 $m^2$. As an example, for one trial, we had an average of 17.5 cm translation between each image and a total of 0.35 $m^2$ area covered.    
When reconstructing the open CAO models, the pipeline described in Section III.A was adapted to accommodate the reduced image count for inference. The NetVLAD retrieval method was replaced with an exhaustive matching strategy, and COLMAP's dense reconstruction pipeline \cite{schoenberger2016sfm} was applied to the sparse point cloud to enhance reconstruction quality with the limited image dataset. This approach, however, sacrifices precise point-to-pixel correspondences, necessitating the use of all segmentation masks in the weighted voting process to generate a segmented point cloud. Additionally, due to the limited amount of data available for fine-tuning the RGB images segmentation network, we adopted the segmentation pipeline described in \cite{jmrr}. It uses the same architecture as in Section~\ref{Sec:SegmentedPC}, but includes a step for automated bounding box generation. In total, running the complete reconstruction and registration pipeline for 20 images takes around 20 to 40 minutes. 

\subsubsection{Model Scaling and Registration}
\begin{figure}[tbp]
\centering
\includegraphics[trim={0cm 0cm 0cm 0cm},clip,width=0.8\columnwidth]{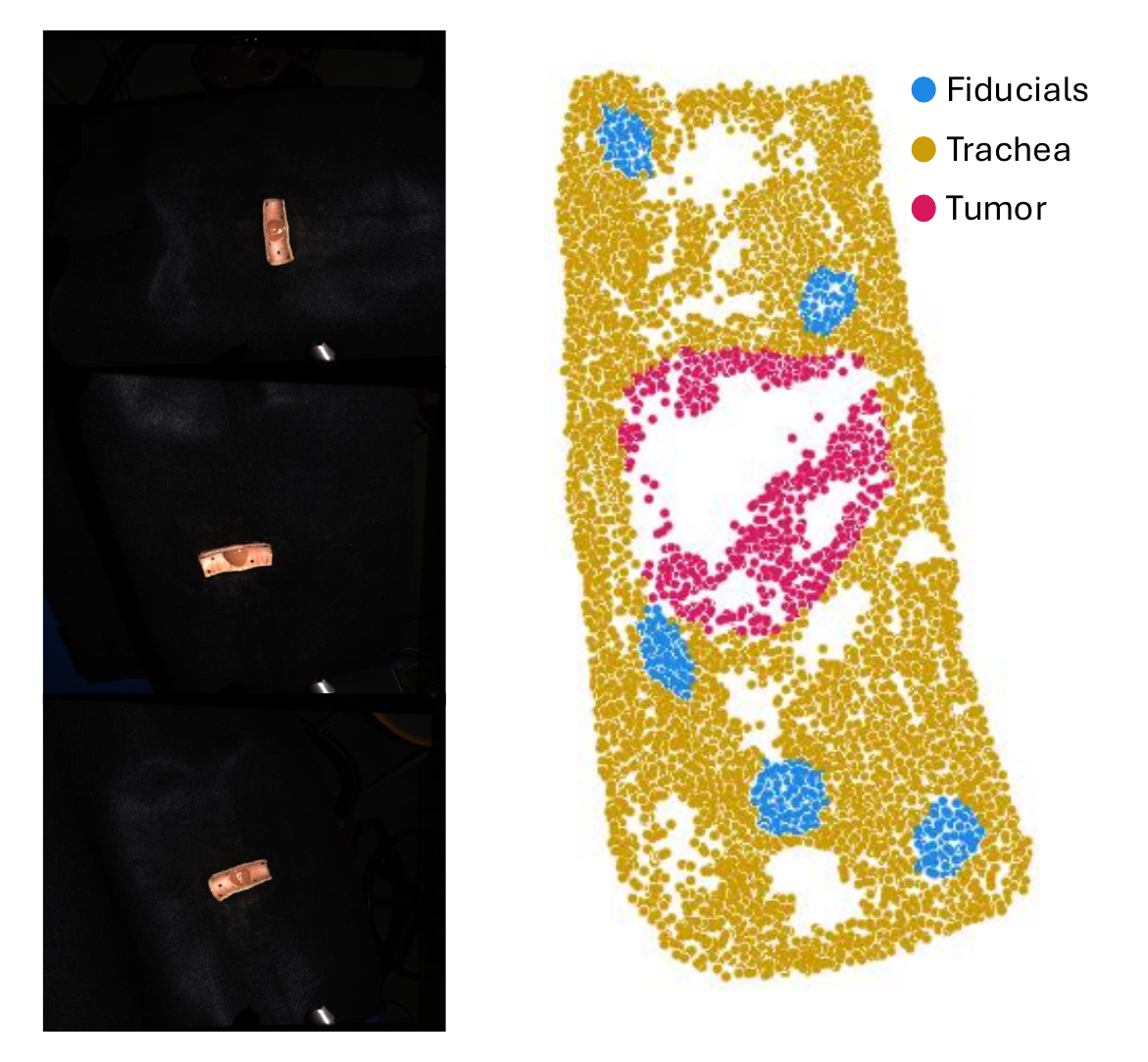}
\caption{Example images from resection experiment setup (left) and resulting segmented point cloud (right).}
\label{OpenCAOReconstruction}
\end{figure}

To facilitate registration between the reconstructed model and the physical specimen, five fiducial markers were cauterized into each trachea at the beginning of the resection procedure, as shown in Fig.~\ref{OpenCAOReconstruction}. The pose of the cautery needle tip at each fiducial was recorded relative to the robot's world frame. The fiducials serve dual purposes: enabling spatial registration, and providing scale reference for the reconstructed trachea model. For fiducial identification in 2D images, segmentation masks were generated using OpenCV adaptive thresholding on the masked trachea images \cite{opencv_library}. We note that significant changes in the environmental conditions may require the change of hyperparameters. The resulting fiducial masks were dilated with a $5\times5$ kernel, followed by contour detection to eliminate areas with circularity values below 0.5. The five largest clusters after point segmentation were selected as fiducial areas (Fig. \ref{OpenCAOReconstruction}). Projecting the fiducial masks onto the point cloud, the centroid of each fiducial cluster is computed to identify the five 3D registration points. Registration and scaling were implemented using Umeyama's approach \cite{umeyama1991least}, which requires corresponding points to maintain consistent ordering. Accordingly, both the reconstructed fiducials and robot needle tip positions were sorted based on their distances to the centroid of their respective sets. The registration accuracy is quantified by calculating the $L_2$ norm between the fiducial points in the scaled and registered point cloud and the corresponding ground truth robot needle tip positions. 

\subsubsection{Resection}
 
Following registration of both the trachea and tumor point clouds to the robot's world frame, resection procedures were conducted according to the protocol established in \cite{jmrr}. A polynomial (poly55) model derived from the trachea point cloud provides a continuous surface representation for planning cut trajectories, while the tumor point cloud defines the start and end coordinates for each incision. Each cut was programmed to maintain a 1\,mm clearance above the trachea surface, advancing in 4\,mm increments into the tumor tissue. During cuts, the electrosurgical tool was maintained at a power of 35\,W. Throughout the procedure, the passive gripper maintained grasp of the tumor, applying consistent tension between cuts to expose the developing incision boundary through a heuristic retraction of constant magnitude and direction. Resection continued incrementally until complete detachment of the tumor from the tracheal surface was achieved. Upon procedure completion, RGB-D data was captured for subsequent char analysis and cut surface evaluation.

Clinically, a successful CAO removal minimizes trachea damage while maximizing the amount of tumor removed. Post-procedure tissue charring analysis was conducted for each model by identifying charred points using a consistent pixel-wise threshold across all specimens. Two quantitative metrics are developed to evaluate tracheal damage: 

\begin{equation}
    \textit{AM} = \frac{\textit{Char Area} - \textit{Tumor Area}}{\textit{Tumor Area}}, 
\label{area}
\end{equation}
\begin{equation}
    \textit{DM} = \frac{\textit{Char Darkness} - \textit{Trachea Darkness}}{\textit{Trachea Darkness}}.
\label{darkening}
\end{equation}

\begin{figure}[tbp]
\centering
\includegraphics[trim={0cm 0cm 0cm 0cm},clip,width=0.95\columnwidth]{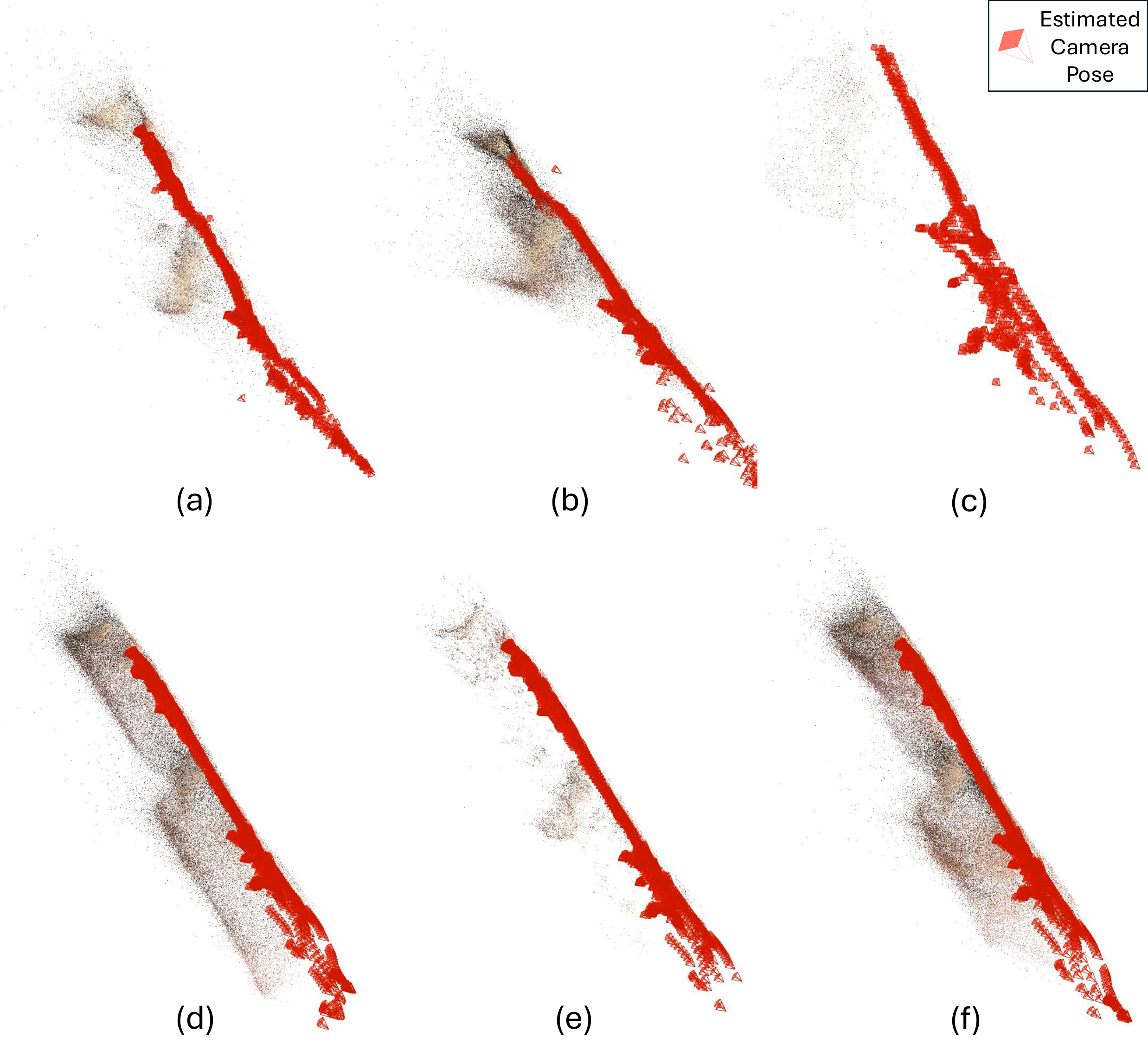}
\caption{Qualitative comparison of six structure from motion pipelines. \textbf{(a)} SuperPoint~\cite{detone2018superpoint} + SuperGlue~\cite{sarlin2020superglue}, \textbf{(b)} SuperPoint + LightGlue~\cite{lindenberger2023lightglue}, \textbf{(c)} SuperPointInLoc~\cite{ taira2018inloc} + LightGlue, \textbf{(d)} ALIKED~\cite{zhao2023aliked} + LightGlue, \textbf{(e)} SIFT NN~\cite{lowe1999object}, \textbf{(f)} DISK~\cite{tyszkiewicz2020disk} + LightGlue}
\label{SfMComparison}
\end{figure}

The Area Margin (AM) (Eq. \ref{area}) measures the extent of charring beyond the original tumor boundaries, and the Darkening Margin (DM) (Eq. \ref{darkening}) quantifies the increased darkness level in charred areas relative to the unaffected tracheal tissue. Consequently, smaller area and darkening margins indicate better resection outcomes.

To quantify tumor removal, a post-procedure cut surface analysis was performed. After each procedure was complete, a point cloud of the tissue was obtained. This point cloud was segmented into 1) the charred area, and 2) the non-charred trachea points. A polynomial surface model was fit to the non-charred trachea point cloud to create a `goal' surface (if the tumor was removed as planned, this surface would align exactly with the charred area). For each point in the charred area, its Z value was compared with the corresponding point in the `goal' surface. By performing this calculation for all charred points in each model, we obtained a cut surface RMSE for each model, where a low RMSE indicates a successful procedure.

\begin{table*}
\centering
\begin{minipage}{0.95\textwidth}
\caption{Average quantitative results for structure from motion comparison.}
\label{Tab:SfMComparisonTable}
\centering
\centering
\begin{tabular*}{0.98\textwidth}{@{\extracolsep{\fill}}c|c|c|c|c|c|c}
\centering
\textbf{Method} &
  \textbf{Reconst. \%} &
  \textbf{\# Points} &
  \textbf{Track Length} &
  \textbf{Observations} &
  \textbf{Reprojection Error} &
  \textbf{Runtime (min)} \\
\hline
SuperPoint + SuperGlue    &62.51 &50025.11  &10.35  &704.05  &1.46  &90.90  \\
\hline
SuperPoint + LightGlue     &77.53  &94897.67  &12.02  &1434.34  &1.62  &106.19  \\
\hline
SuperPointInLoc + LightGlue &48.15  &4379.33  &11.44  &109.88  &1.40  &19.25  \\
\hline
ALIKED + LightGlue          &\textbf{99.66}  &72225.67  &19.63  &1644.74  &1.68  &113.33  \\
\hline
SIFT NN                     &58.57  &7098.78  &17.38  &198.03  &\textbf{1.00}  &\textbf{9.89}  \\
\hline
DISK + LightGlue            &92.13  &\textbf{116584.89}  &\textbf{25.73}  &\textbf{3457.76}  &1.73  &243.21 
\end{tabular*}
\end{minipage}
\end{table*}

\section{Results}
\label{Results}

\subsection{CAO Model Reconstruction}
\subsubsection{SfM Results}
Based on the quantitative results shown in Table~\ref{Tab:SfMComparisonTable}, we observe that the highest reconstruction rate (number of total images/number of images used in reconstruction) is in ALIKED~\cite{zhao2023aliked} + LightGlue~\cite{lindenberger2023lightglue} pipeline. However, in metrics such as the number of points in the reconstructed point cloud which indicates the density, track length and observations per image, DISK~\cite{tyszkiewicz2020disk} + LightGlue~\cite{lindenberger2023lightglue} outperforms the others. SIFT~\cite{lowe1999object} NN pipeline offers faster reconstructions and a low reprojection error, however, this network results in very sparse reconstructions as can also be seen in qualitative results (Fig.~\ref{SfMComparison}). 

In both qualitative and quantitative results, ALIKED and DISK perform similarly and both are better than the other methods. Since the DISK pipeline produces significantly denser reconstructions with a higher number of points, we proceed with it.

\subsubsection{Segmented Point Clouds}
Our analysis reveals that when we reproject the tumor point clouds to the ground truth labels, 91.74\% of points fall within the segmentation masks. Per-image average is 91.95\% ($\pm$ 3.62\%). This indicates a high precision of point cloud segmentation. DICE similarity coefficient between 2D masks manually labeled and generated by our segmentation model is 92.69\% ($\pm$ 10.14\%).

\subsection{Autonomous Robotic Resection}
We achieve 0.41 $\pm$ 0.16\,mm registration accuracy between the fiducials in the point cloud and recorded positions of end effector for all five samples.

\begin{figure}[htbp]
\includegraphics[trim={0cm 0cm 0cm 0cm},clip,width=1\columnwidth]{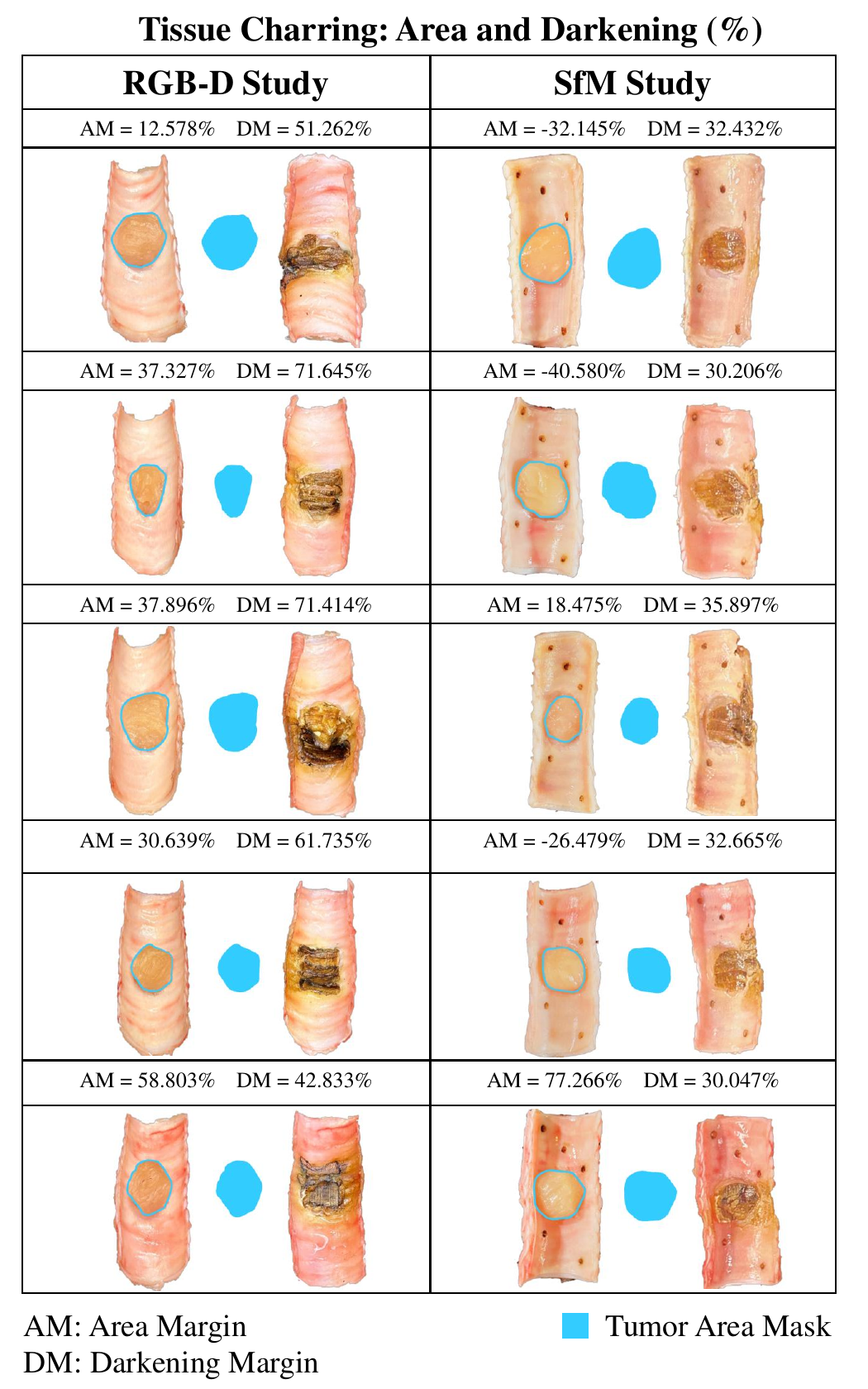}
\caption{Comparison of tissue charring metrics (area margin and darkening margin) between the RGB-D-guided resection in \cite{jmrr} and the SfM-guided resection presented in this work. Tumor area margin measures the extent of charring beyond the original tumor area, given in blue.}
\label{CharAnalysis}
\end{figure}

\begin{figure}
\includegraphics[trim={0cm 0cm 0cm 0cm},clip,width=1\columnwidth]{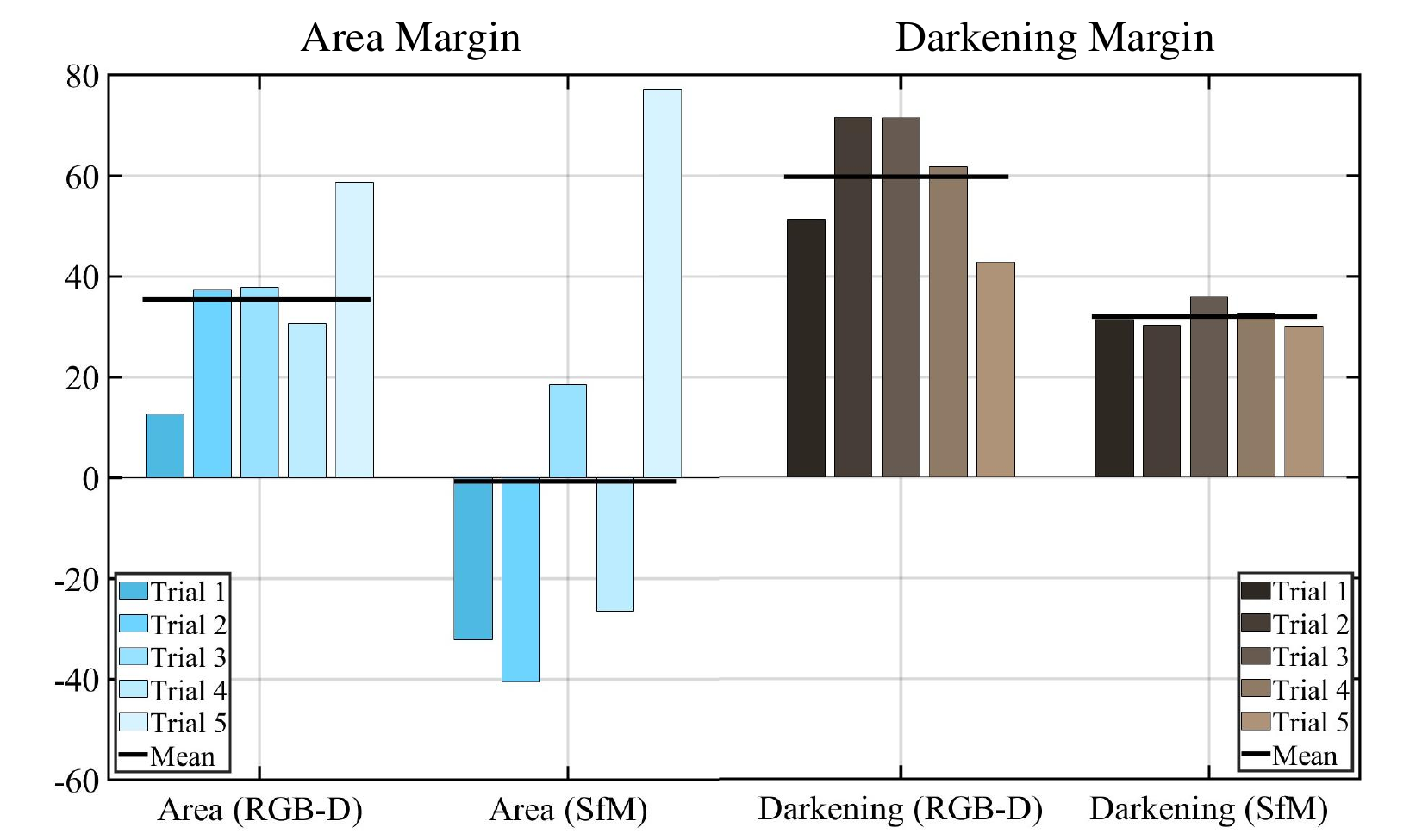}
\caption{Comparative results of area and darkening margins for the RGB-D study \cite{jmrr} and the SfM study presented in this work.}
\label{AMDM}
\end{figure}

As shown qualitatively in Fig. \ref{CharAnalysis} and quantitatively in Fig. \ref{AMDM}, the study conducted in this work using SfM yielded a significant reduction in mean AM and DM compared to our prior study \cite{jmrr} which used RGB-D guidance, indicating an observable reduction in trachea damage. 

\begin{figure}[tbp]
\includegraphics[trim={0cm 0cm 0cm 0cm},clip,width=1\columnwidth]{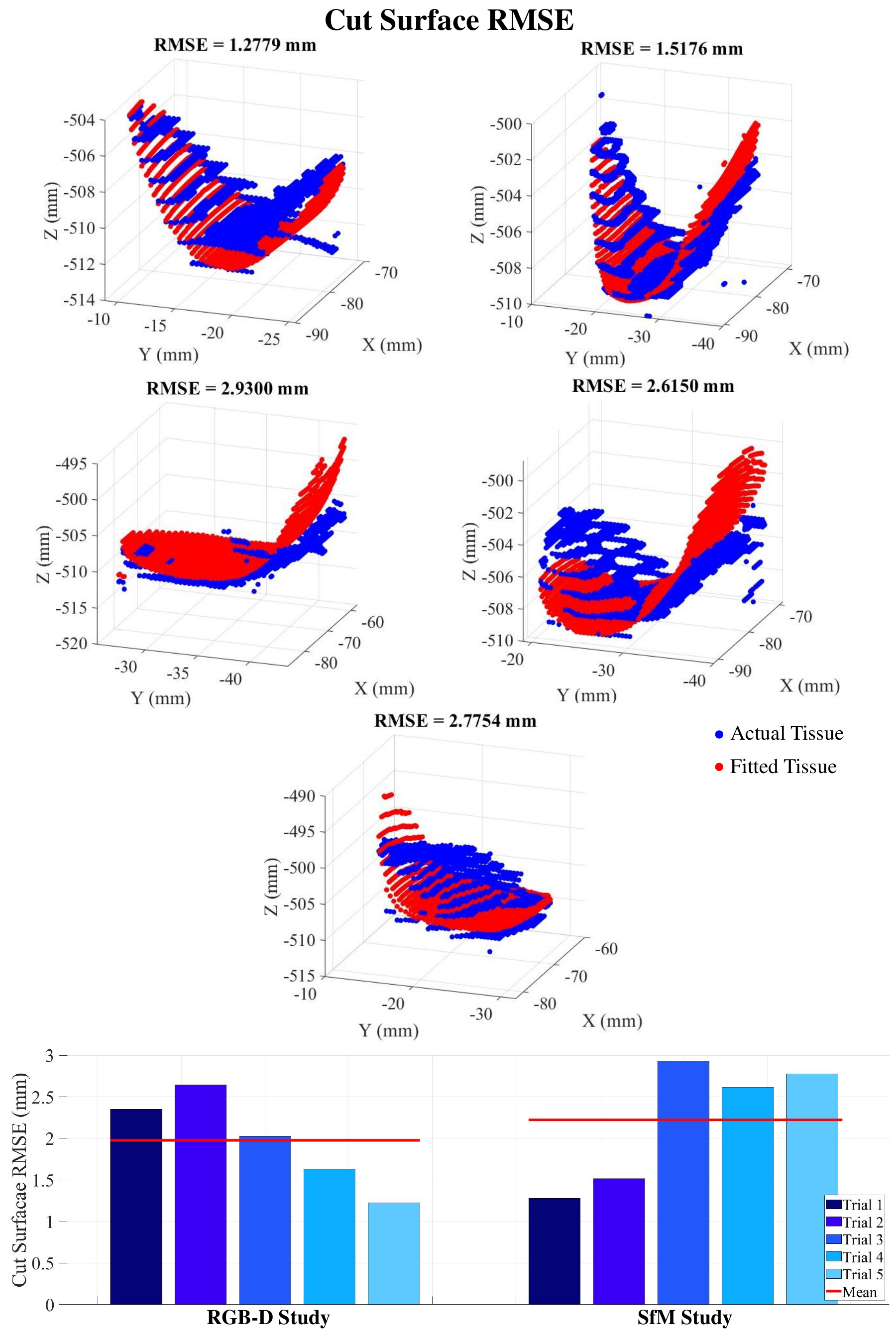}
\caption{Visualization of the charred tissue surface ('Actual Tissue') versus the goal surface ('Fitted Tissue'), for computation of cut surface RMSE in all five models. Comparative results of cut surface RMSE for the RGB-D study \cite{jmrr} and the SfM study presented in this work. Low RMSE is desirable.}
\label{CutRMSE}
\end{figure}

SfM method in this study yielded a comparable mean cut surface RMSE compared to our prior study \cite{jmrr} which used RGB-D guidance (1.976\,mm for RGB-D and 2.223\,mm for SfM) (Fig. \ref{CutRMSE}). 

\section{Discussion}
\label{Discussion}
Quantitative and qualitative results show the feasibility of acquiring segmented point clouds through reconstruction. Tests on the downstream task show that RGB-based proposed pipeline can indeed replace the bulkier and more expensive RGB-D cameras.

Regarding tissue charring, the SfM study showed a significant reduction in AM and DM compared to the RGB-D approach. These results show that our SfM approach causes an observable reduction in trachea damage, possibly indicating that our fiducial-based registration outperforms an RGB-D hand-eye calibration in spatial accuracy. We hypothesize that this increase in accuracy is attributable to the fusion of 20 different camera views for reconstruction and segmentation, rather than a single partial-view snapshot as used in the RGB-D approach. Also, the fiducial markers are concentrated in the critical region of the robot's workspace (the tissue location). This allows for more precise localization of the tissue, resulting in more accurate tracking of cut trajectories, ensuring less trachea damage. However, we must note that this study is a comparison of SfM to a single-view RGB-D snapshot, and not a fusion of multiple RGB-D views. Better RGB-D performance could be obtained by stitching together depth images from multiple views into a single map. 

The cut surface RMSE results, along with the charring analyses, demonstrate that the SfM approach provides visualization quality comparable to the RGB-D method, leading to similarly successful resection trajectories and overall procedure outcomes. This suggests that the RGB-D approach in our autonomous resection workflow can be replaced with this SfM-based method, enabling vision guidance using a monocular camera. This modification enables our workflow to be integrated into minimally invasive monocular robotic systems designed to operate within the constrained lumen of the trachea. 

However, crucial considerations arise with SfM when considering the dynamic nature of in-vivo tissues. With an RGB-D camera, a full updated map can be generated in a fraction of a second. The full SfM vision pipeline used in this study had a time scale of minutes. In the future, to address tissue motions like breathing and deformation, this SfM vision pipeline must be streamlined such that it can be completed in a matter of seconds or sub-seconds (such as during a breath hold), or should be integrated with a deformation model. Additional automation, such as automated fiducial touchpoints or camera movement, and a more optimized runtime could further this work in pursuit of that goal. 

Performance of the pipeline is not limited by the performance of individual components. Due to plug-and-play design of the algorithm, components can be replaced with better performing counterparts as the technology progresses. This also allows easier adaptation to different downstream tasks and surgeries.

\section{Conclusion}
\label{Conclusion}
To conclude, in this paper we present a pipeline to create segmented point clouds from 2D image sequences to guide CAO surgeries. We comprehensively and comparatively evaluated SfM algorithms in a dataset consisting of ex-vivo CAO images to determine the optimal method for CAO. We tested our pipeline in a tumor resection task and achieved results comparable to, or even better than, the results obtained with RGB-D cameras. Our future work will focus on the integration of real-time 3D reconstruction algorithms, and tests on minimally invasive surgical robots as well as different surgical applications.

\bibliographystyle{IEEEtran}
\bibliography{bibliography}

\end{document}